# Building predictive models of healthcare costs with open healthcare data

A. Ravishankar Rao, Fellow, IEEE
Fairleigh Dickinson Univ., NJ, USA
raviraodr@gmail.com

Subrata Garai
IT Software Engineer
subratagarai@gmail.com

Soumyabrata Dey, PhD.
Machine Learning Researcher
soumyabrata.dey@gmail.com

Hang Peng
Machine Learning Engineer
hanspenghang@gmail.com

*Abstract*

**Due to rapidly rising healthcare costs worldwide, there is significant interest in controlling them. An important aspect concerns price transparency, as preliminary efforts have demonstrated that patients will shop for lower costs, driving efficiency. This requires the data to be made available, and models that can predict healthcare costs for a wide range of patient demographics and conditions.**

**We present an approach to this problem by developing a predictive model using machine-learning techniques. We analyzed de-identified patient data from New York State SPARCS (statewide planning and research cooperative system), consisting of 2.3 million records in 2016. We built models to predict costs from patient diagnoses and demographics. We investigated two model classes consisting of sparse regression and decision trees. We obtained the best performance by using a decision tree with depth 10. We obtained an $R^2$ value of 0.76 which is better than the values reported in the literature for similar problems.**

## I. INTRODUCTION AND MOTIVATION

Considerable research is being conducted to minimize costs and maximize efficiency in healthcare [1]. In the US, it has been very difficult to obtain data related to healthcare costs as this is closely guarded by hospitals, insurance companies and providers. Hence, the current administration in the US is aiming to increase price transparency [2]. There are available sources such as the New York State SPARCS dataset [3]. There is sufficient information in this dataset for researchers to start formulating questions, analyzing the existing data, creating models, and exploring promising directions.

We focus on two three main research questions: (1) Can we reliably predict the cost of a procedure given its diagnosis code, severity, age and demographic information about a patient? (2) What are the most important factors that influence the cost of healthcare procedures? (3) How can we facilitate the reproducibility of our research methodology? We tackle these through a unified approach.

## II. BACKGROUND AND RELATED WORK

Earlier work by Rao et al. includes the analysis [4-7] of open health data from primarily the Center for Medicare and Medicaid Services (CMS) [8]. Prior explorations in healthcare cost prediction include those by Sushmita et al. [9], Bertsimas et al. [10], Cumming et al. [11], Zikos [12] and others [13-17]. Using New York SPARCS data leads to an understanding of the reimbursement limits that insurance companies may place on medical procedures [13]. Morid et al. [14] provide a survey of supervised classification techniques that have been used for predicting healthcare costs.

## III. DESIGN

In our earlier papers, we presented the architecture of an open-source system for the analysis of open health data[5, 15].

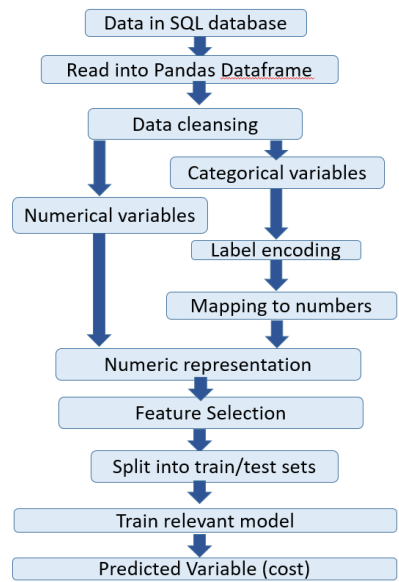

**Figure 1:** The sequential steps used for model building.

**We extend our earlier architecture to include modules on predictive modeling as shown in**

Figure 1. We use a Python-based solution with the following components: Python Pandas, Scikit-Learn [18] and Matplotlib [5].

## IV. METHODS

We examined approximately 2.3 million patient records for the year 2016 from SPARCS [3]. We split the data into a holdout test dataset and a training dataset. The holdout test data comprised 50%, and the training dataset was 50% of the samples. We evaluated the following two model classes: sparse regression and decision trees. We used the $R^2$ metric for performance evaluation. We calculated the final model $R^2$ performance metric for the holdout test set only.

## V. RESULTS

We used three measures of feature importance: the chi square measure, the mutual information measure and XGBoost feature importance measure. We chose the union of the top 5 features identified by each of these techniques to create a list of 11 features for analysis. These included the following: Operating cert. number, length of stay, CCS Diagnosis code, APR DRG Code. Payment typology, Ethnicity, APR Medical surgical description, APR Risk of mortality, gender, emergency dept. indicator and APR Severity of illness code.

We used the DecisionTreeRegressor algorithm in Scikit-Learn [18]. We used cross validation to determine the maximum depth of the tree, which was 10. We also used Linear Regression, Ridge Regression, Lasso and ElasticNet. We summarize the best performing methods in Table 1.

| Model type | RMS Error | $R^2$ |
|---|---|---|
| Ridge regression | $13,239 | 0.71 |
| LassoLarsIC-AIC | $13,190 | 0.72 |
| LassoLarsIC-BIC | $13,407 | 0.71 |
| Decision Tree | $12,652 | 0.76 |

**Table 1:** A comparison of the root mean squared error and the $R^2$ metric for the best models we investigated. The shaded rows represent performance improvements with respect to earlier work [16].

In Figure 2 we visualize the performance of the best model in Table 1 (Decision Tree Regression) by comparing the actual costs (x-axis) versus the predicted costs (y-axis). The diagonal blue line represents an ideal model, where the actual and predicted costs are equal.

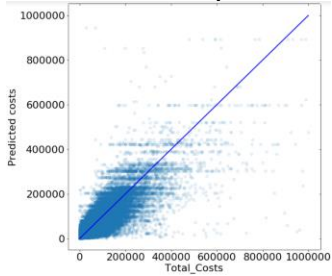

**Figure 2:** Scatter plot of predicted costs with the decision tree vs. total costs. All the points in the test set are shown. The Pearson correlation is 0.76, p < 0.001.

## VI. DISCUSSION

The correlation values in Figure 2 and are high to moderately high, and so are the $R^2$ values in Table 1. This indicates success of our predictive modeling effort. In Table 2, we compare efforts by several authors in the literature. We caution that this comparison is limited as each author uses a different dataset, which is generally not accessible. We have included an estimate of the size of the data that the models were developed and tested on.

| Author | Type of Model | Size of Data | Patient Age | $R^2$ |
|---|---|---|---|---|
| Evers, 2002 | Multiple Regression | 731 | ~75 (avg.) | 0.61 |
| Cumming, 2002 | Multivariate linear regression | 749,145 | All | 0.198 |
| Bertsimas, 2008 | Classification Trees | 838,242 | All | 0.2 |
| Zikos, 2016 | Multiple Regression | 1 M | > 65 | 0.66 |
| 2020 | LassoLarsIC-AIC | 2.3 million | All | 0.72 |
| 2020 | Decision Tree Regression | 2.3 million | All | 0.76 |

**Table 2:** Compares $R^2$ values in the literature, sorted by date. The values in the current paper are reported as "2020". Shaded rows represent improvements over the previous best $R^2$ values.

Table 2 demonstrates that two of the methods investigated in this paper have a superior performance relative to the previous efforts.

The use of our techniques along with open health data could provide citizens with valuable personalized medical cost information that can help them predict procedure costs. Our predictive models can help policy makers and administrators. The transparency of our entire approach, including data, algorithms, and code supports trust [19], and the reproducibility of scientific experiments [20]. Our framework is freely available at www.github.com/fdudatamining/.

## VII. CONCLUSION

We presented and evaluated predictive models for healthcare costs using machine learning techniques. We used open data provided by the New York State SPARCS program, consisting of approximately 2.3 million de-identified patient records in 2016.

We investigated two main classes of techniques, including sparse regression and regression trees We obtained the best performance with $R^2 = 0.76$ by using a decision tree with depth 10. This improves upon the previous best result of $R^2 = 0.71$ reported in this class of problems in the literature. The most significant aspect of our result is that patients can readily predict the costs of a wide range medical procedures they may undergo. This is currently not easy to do in the US. Policy makers can use our model to predict healthcare costs for planning purposes.